\documentclass[12pt]{article}
\usepackage{times,apalike,url}
\title{Computational Phonology}
\author{Steven Bird}
\date{University of Pennsylvania}
\begin{document}
\maketitle

Phonology, as it is practiced, is deeply computational.
Phonological analysis is data-intensive and the resulting
models are nothing other than specialized data structures and algorithms.
In the past, phonological computation -- managing data and developing
analyses -- was done manually with pencil and paper.  Increasingly,
with the proliferation of affordable computers, IPA fonts and drawing software,
phonologists are seeking to move their computation work online.
{\em Computational Phonology} provides the theoretical and technological
framework for this migration, building on methodologies and tools from
computational linguistics.  This piece consists of an {\em apology} for
computational phonology, a history, and an overview of current research.

{\bf Documentation and Description.}
Phonological data is of essentially three types:
texts, wordlists and paradigms.
A text is any phonetically transcribed narrative or conversation.
A wordlist is any compilation of linguistic forms which can be uttered
in isolation, with information about pronunciation and meaning.
A paradigm is broadly construed to mean any
tabulation of words or phrases which illustrates
contrasts and systematic variation.
Any of these data types may be {\em annotated} with more abstract
information originating from a phonological theory, such
as syllable boundaries, stress marks and prosodic structure.
Additionally, any of these data types may be associated with
recordings of audio, video or physiological signals.
Digitizing this documentation and description brings all the
different media types together, makes the cross-links
navigable, and opens up many new possibilities for
management, access and preservation.

{\bf Exploration and Analysis.}
The data types described above are closely interconnected in
phonological practice.
For instance, the discovery of a new word in a text
may require an update to the lexicon and
the construction of a new paradigm (e.g. to correctly
classify the word).  Fresh insights may lead to new annotations
and further elicitation, closing the loop in this perpetual,
exploratory process.
Phonological analysis typically involves defining a formal model,
systematically testing it against data, and comparing it with other models.
(In some cases, the model may be incorporated into a software system, e.g. for
generating natural intonation in a text-to-speech system.)
In this exploration and analysis --
sorting, searching, tabulating, defining, testing and comparing --
the principal task is computational.

Perhaps the earliest work in computational phonology was Bobrow and
Fraser's {\em Phonological Rule Tester} \cite{BobrowFraser68}, an
implementation of SPE designed to ``alleviate the problem of rule
evaluation.''  Shortly afterwards Johnson showed that, while SPE rules
resemble general rewriting systems at the top of the
Chomsky hierarchy, the way SPE rules are used in practice only requires
finite state power \cite{Johnson72}.
Independently, Kaplan and Kay discovered the connections between SPE
grammars and finite state transducers in the 70's and 80's, and laid down
a complete algebraic foundation (ultimately reported in
\cite{KaplanKay94}).  Significant implementations followed, including
\cite{Koskenniemi83b,BeesleyKarttunen02}.  Attempts to apply finite
state devices to Autosegmental Phonology have largely foundered,
but applications to Optimality Theory are thriving.

While finite-state phonology fixated on SPE, generative phonology continued
its rapid evolution.  The discovery of rule ``conspiracies''
\cite{Kisseberth70} and the abstractness controversy \cite{Koutsoudas74},
lead to calls for the reintroduction of {\em surface structure
constraints}.  Many theories arose from the fallout; most notable
for its computational ramifications was
Montague Phonology \cite{Wheeler81}.  This model adapted new lexicalist
formalisms from syntax and semantics, providing a {\em declarative}
(as opposed to {\em procedural}) account of phonological well-formedness,
and providing the first computational account of underspecification (where the
phonological content of a lexical entry is incompletely specified,
to be filled in during a derivation).  From these beginnings, {\em Declarative
Phonology} was born, and subsequent work provided a mathematical foundation
in first-order logic \cite{Bird95} and phonetic interpretation with links
to Firthian prosodic analysis and speech synthesis \cite{Coleman97},
with implementations generally in the Prolog programming language.

A third major strand of development, complementing the finite state and
declarative models, is best characterized as statistical.  It seeks to
apply neural networks, information theory, and weighted automata
in the automatic discovery of phonological information.
Gasser trained a recurrent neural network to recognize syllables and to repair
ill-formed syllables \cite{Gasser92a}.
Ellison showed how a technique from information theory called
MDL -- {\em minimum description length} -- could be applied to
automatically identify syllable boundaries in phonemically transcribed
texts \cite{Ellison92b}.
Many researchers apply Markov models (a kind of weighted automata) in
speech recognition, mapping speech recordings to phonetic transcriptions
and thence to orthographic words, using large, phonetically annotated
corpora as training data (e.g. TIMIT \cite{TIMIT86}).

Four key areas of ongoing research in computational phonology are in
Optimality Theory, automatic learning, interfaces to grammar and
phonetics, and supporting phonological description in the field.
Comprehensive references to online research papers in this areas may
be found on the SIGPHON website.

Computational phonology is generating sophisticated and rigorous ways for
creating, exploring and disseminating multidimensional phonological
information, encompassing primary recordings, texts, wordlists, paradigms,
theories and analyses.  As phonologists adopt the computational methods
described above, extending and adapting them as needed, the consequences
for the discipline will be increased accessibility, accountability, and
stability of empirical research.

{\bf Resources.}
The Association for Computational Linguistics
(ACL) has a special interest group in computational phonology (SIGPHON) with
a homepage at \url{http://www.cogsci.ed.ac.uk/sigphon/}.
The website contains online proceedings for SIGPHON workshops
and information about relevant books, dissertations and articles.
A special issue of {\it Computational Linguistics} devoted to
computational phonology was published in 1994 \cite{Bird94}.

\raggedright

\end{document}